\documentclass{article}
\usepackage[T1]{fontenc}
\usepackage[margin=1in]{geometry}
\usepackage{amsmath,amssymb}
\usepackage{graphicx}
\usepackage{booktabs}
\usepackage{array}
\usepackage{subcaption}
\usepackage{caption}
\usepackage{enumitem}
\usepackage{xcolor}
\usepackage{hyperref}
\usepackage{float}
\usepackage{placeins}
\hypersetup{colorlinks=true, linkcolor=blue, urlcolor=blue, citecolor=blue}

\newcommand{\projectname}{QFireNet}

\title{\projectname: A Quantum-Enhanced U-Net for Wildfire Segmentation\\ from Sentinel-2 Imagery}

\author{
Jaiman Munshi\thanks{Corresponding author: \texttt{jmunshi@umd.edu}.} \and Tanvi Tewary \and Sawyer Bloom
\and Aidan Chu \and Chetan Maviti \and Kyon Winston-Bey \and Harshit Badjatia \and Farhan Kittur \and Vardhan Madhavarapu \and Varun Kota \and Joshua Kwon \and Nazia Rangwala-Vohra \and Franz Klein\thanks{Corresponding author (advisor): \texttt{fklein@umd.edu}.}
\\[0.5em]
\normalsize IonQ Team, App Dev Club, University of Maryland, College Park
}

\date{}

\begin{document}

\maketitle

\begin{abstract}
Wildfire detection from satellite imagery is a semantic image segmentation problem that has proven to be difficult due to challenges such as class imbalance, feature complexity, and atmospheric interference. In this paper, we build on the foundational U-Net image segmentation model to develop a quantum-hybrid solution in hopes of more effectively modeling the high-dimensional spectral feature space of the Sen2Fire dataset. We inject a variational quantum circuit in the bottleneck portion of U-Net, specifically the QuFeX and QB-Net ansatzes.  We test a classical Feature Pyramid Network (FPN) for further comparative analysis of the model, and we also explore classical improvements to the U-Net model and its training process, including a compression of parameters, alternative loss functions, and uniform mixing of input data. Our primary finding is that under matched conditions, both QB-Net (with an $F_1$ score of 31.18) and QuFeX ($F_1 = 30.79$) outperformed the classical U-Net baseline results ($F_1 = 28.71$). Additionally, the classical FPN achieved a comparable score of 31.13. A crucial finding was that data mixing removed a significant domain shift between the geographically-separated train and test sets, which boosted the classical FPN $F_1$ score to 39.76. We validate the architecture's robustness and generalizability to the wildfire detection problem via cross-dataset transfer on the California Burned Areas (CaBuAr) dataset. Overall, we find that quantum machine learning has potential to provide an advantage in the problem of wildfire image segmentation, and further experiments will continue to validate and expand upon this finding.
\end{abstract}

\section{Introduction}

The increasing frequency and intensity of global wildfires make manual fire monitoring inefficient and unreliable, leading to public safety and environmental risks. Automated systems capable of processing large streams of satellite imagery with high precision and recall are critical to preserving ecosystems and preventing destruction.

Despite the urgent need for a fully automated wildfire detection system, developing one presents numerous technical challenges:

\begin{enumerate}
    \item \textbf{Class Imbalance:} Since fire events are rare and typically make up a tiny fraction of pixels in all satellite images, standard machine learning models will often achieve ``high accuracy'' by simply predicting `non-fire' for every pixel, leading to actual fires being missed.
    \item \textbf{Spectral Complexity:} Wildfires tend to vary greatly depending on their intensity, type of burn, and the surrounding environment.
    \item \textbf{Atmospheric Interference:} Smoke, clouds, and aerosols can lead to false positive detections or obfuscate detection of real fires.
    \item \textbf{Multi-Scale Nature:} Fires can range from tiny ignition points to large conflagrations, requiring a model that can capture details and extract patterns on both small and large scales.
\end{enumerate}

The volume, dimensionality, and irregularity of satellite wildfire data exceed what hand-designed thresholds or rule-based algorithms can robustly handle. Each revisit by the Sentinel-2 satellite produces gigabytes of multispectral imagery per study area, and any individual fire signature depends on a non-linear interaction between spectral bands, surrounding land cover, and atmospheric conditions. Spectral indices like Normalized Burn Ratio (NBR) and Normalized Difference Vegetation Index (NDVI) capture only a small slice of this interaction and saturate quickly under smoke, cloud cover, or unusual fuel types. 

Our training dataset, Sen2Fire~\cite{sen2fire}, relies on multiple satellite technologies combining imagery and atmospheric data to clearly picture the Earth's ground. The Sentinel-2 Mission provides continuous monitoring via its Multispectral Imager (MSI) aboard the Sentinel-2A and Sentinel-2B satellites, offering a five-day revisit time. It captures 13 distinct spectral bands across the electromagnetic spectrum at spatial resolutions ranging from 10 to 60 meters. These bands are critical because high-temperature targets emit primarily in the short-wave infrared region of the spectrum. Additionally, we utilize the Aerosol Index from the Sentinel-5P satellite. Because wildfires produce significant amounts of smoke, aerosol optical depth data provides crucial atmospheric correction, allowing the model to accurately identify features under varying atmospheric conditions. To train the model, we rely on ground truth labels derived from the MOD14A1 V6.1 global daily fire product to manually annotate the binary fire masks.

Early methods of fire detection relied on simple threshold-based algorithms that looked at single spectral bands such as NBR and NDVI. These early approaches tended to suffer from high false-positive rates and could not adapt to varying environments. Machine learning, and convolutional neural networks in particular, address this gap by learning the relevant feature interactions directly from labeled data. Wildfire image segmentation models like U-Net \cite{unet} combine Short-Wave Infrared (SWIR) reflectance, aerosol concentration, and local spatial context into a per-pixel fire probability without a human specifying what those rules should be, and it generalizes to scenes the analyst never explicitly considered.

Even within the deep learning family, however, wildfire detection remains a difficult problem. The class imbalance is high (less than 3\% of pixels in Sen2Fire are fire pixels), the input is high-dimensional (12 spectral bands plus aerosol), and the most informative features sit at the deepest, most compressed point of the machine learning model, where a classical convolution must summarize the entire scene through a small number of channels. This is where quantum machine learning can provide an advantage. Variational quantum circuits operate in a Hilbert space whose dimension grows exponentially with qubit count, providing a high-capacity feature space for compressed, correlated inputs at relatively few trainable parameters. Hybrid quantum-classical architectures replace selected layers of a classical network, typically at the bottleneck, with such circuits, with the goal of capturing complex correlations between features that an equivalent classical layer cannot represent. While near-term quantum hardware remains noisy and limited in scale, simulator-based experiments allow us to test whether a quantum bottleneck offers a useful inductive bias for tasks like wildfire segmentation, where the signal of interest is sparse and the spectral feature space is rich.

\paragraph{Contributions.} This paper makes the following contributions:
\begin{itemize}[leftmargin=2em,topsep=2pt,itemsep=2pt]
    \item We implement and compare two variational quantum bottleneck ansatzes --- QuFeX and QB-Net --- as drop-in replacements for the deepest U-Net convolution. Under matched conditions on Sen2Fire, both exceed a classical baseline (QB-Net Fire $F_1 = 31.18$, QuFeX $30.79$, vs.\ the baseline $28.71$).
    \item We introduce a compact U-Net architecture that contains $43.73\%$ less parameters with negligible loss in $F_1$ score on the test set.
    \item We identify and quantify a severe train/test domain shift induced by Sen2Fire's scene-based split (a $5.3\times$ difference in aerosol), and show it sets a substantial performance ceiling under the official split: a uniform random re-split lifts the strongest classical model by $+8.75$ Fire $F_1$.
    \item We show that adopting a Feature Pyramid Network (FPN) decoder and dropping the heavily-shifted aerosol channel (Mode 4) improves the classical baseline to an $F_1$ score of $31.13$, demonstrating a classical mitigation strategy for the domain shift.
    \item We demonstrate cross-dataset portability by transferring the pipeline to the California Burned Areas (CaBuAr) dataset.
    \item We report informative negative results: an initial amplitude-encoding quantum circuit bottleneck produced NaN gradients, and a fire-aware data oversampler collapsed the model's precision.
\end{itemize}

\section{Related Work}

\paragraph{Wildfire segmentation from satellite imagery.} Early approaches to fire detection from multispectral imagery relied on hand-engineered spectral indices such as NBR and NDVI thresholded at fixed values, which suffer from high false-positive rates under varying atmospheric and surface conditions. Convolutional segmentation networks (the U-Net~\cite{unet} in particular) replaced these with end-to-end learned models that combine reflectance and spatial context per pixel. The Sen2Fire benchmark~\cite{sen2fire} dataset on which we run our experiments provides one of the few publicly released Sentinel-2 wildfire segmentation datasets with manually annotated fire masks and a fixed scene-based split; the accompanying baseline U-Net at 17.3\,M parameters defines the current published Fire $F_1$ on this protocol. CaBuAr~\cite{cabuar}, which we use for our cross-dataset transfer evaluation in Section~\ref{sec:cabuar}, extends the benchmark space to California burned areas with a 5-fold cross-validation protocol but is provided at scene granularity in HDF5 and lacks an aerosol channel. Feature Pyramid Networks~\cite{fpn} originate in object detection but transfer naturally to segmentation by replacing the U-Net's symmetric decoder with a top-down pyramid. We use this architectural variant as a strong classical control in Section~\ref{sec:fpn-method}.

\paragraph{Hybrid quantum-classical neural networks.} Recent studies replace selected classical layers of a deep neural network (typically convolutions or dense bottlenecks) with parameterized quantum circuits, motivated by the exponentially large Hilbert space these circuits operate in. Two recent designs targeting the U-Net bottleneck inform our work directly. Qu-Net's QuFeX module~\cite{qufex} splits the qubit register and applies $U_1$ ($R_X{+}R_Z$) and $U_2$ ($R_X{+}R_Y$) blocks with a CZ entangling ring; QB-Net~\cite{qbnet} encodes inputs with $R_Y$ embedding, applies a $U_3$ rotation on every qubit, and entangles via an open CNOT ladder. We adopt both circuits as drop-in bottleneck replacements inside an otherwise identical full-width U-Net, which isolates any classical/quantum performance difference to the circuit itself rather than to backbone changes. A known limitation in this family of architectures is gradient vanishing on deep or wide circuits, known as the barren-plateau phenomenon~\cite{barren-plateaus}. This constrains practical circuit depth on near-term hardware and motivates the modest qubit counts ($n \le 8$) and layer depths ($L \le 2$) we use in our experiments.

\paragraph{Class imbalance and regularization in dense prediction.} Fire pixels comprise under $6\%$ of the Sen2Fire test set, making this a highly imbalanced dataset where the model must learn from very few positive examples. We compare weighted cross-entropy, Dice-augmented losses, and focal variants as standard mitigations in Section~\ref{sec:lossablation}. On the regularization side, we apply both the MixUp~\cite{mixup} and CutMix~\cite{cutmix} sample-mixing strategies, which are now standard ingredients in modern segmentation training pipelines, inside the FPN variant of our backbone described in Section~\ref{sec:fpn-method}.

\section{Sen2Fire Benchmark Dataset}

We train and evaluate \projectname{} on the Sen2Fire benchmark~\cite{sen2fire}, a publicly released Sentinel-2 wildfire segmentation dataset covering the 2019--2020 Australian bushfire season. The dataset comprises 2{,}466 image patches of $512\times512$ pixels, partitioned into four contiguous scenes summarized in Table~\ref{tab:scenes}. Each patch carries (i) a 12-band Sentinel-2 multispectral image, (ii) a single-band aerosol layer derived from Sentinel-5P, and (iii) a binary fire mask manually annotated by domain experts using the MOD14A1 V6.1 daily fire product as a reference signal. Adjacent patches within a scene overlap by 128 pixels along both spatial axes, so a set of patches is reconstructible into a full scene at inference time without seams.

\begin{table}[H]
    \centering
    \begin{tabular}{lcccc}
        \toprule
        Scene & Role & Patch grid (rows $\times$ cols) & Patches & Approx.\ fire-pixel rate \\
        \midrule
        Scene 1 & Train & $32 \times 27$ & 864 & 2--3\% \\
        Scene 2 & Train & $22 \times 27$ & 594 & 2--3\% \\
        Scene 3 & Validation & $14 \times 36$ & 504 & $\approx 3.6$\% \\
        Scene 4 & Test & $21 \times 24$ & 504 & $\approx 5.8$\% \\
        \bottomrule
    \end{tabular}
    \caption{Sen2Fire scene composition and split. The official partition is scene-based: training uses scenes 1 and 2, validation uses scene 3, and testing uses scene 4. Due to the fact that the splits are geographically and temporally distinct, the test set differs substantially from training in aerosol load and fire-pixel density (see Section~\ref{sec:domainshift}).}
    \label{tab:scenes}
\end{table}

\begin{figure}[H]
    \centering
    \begin{subfigure}[t]{0.48\textwidth}
        \centering
        \includegraphics[width=\textwidth]{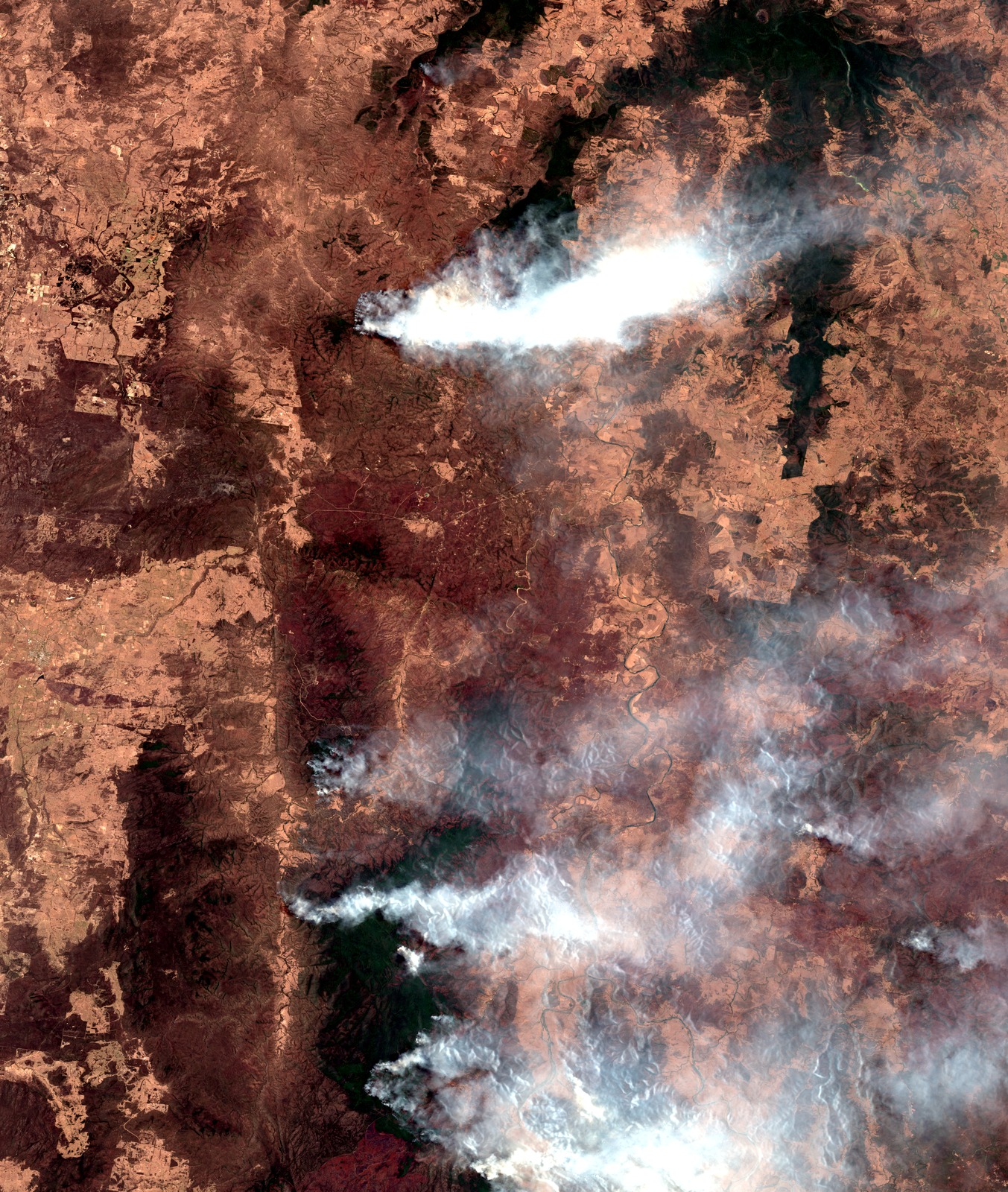}
        \caption{Scene 1 (training).}
        \label{fig:scene1-rgb}
    \end{subfigure}
    \hfill
    \begin{subfigure}[t]{0.48\textwidth}
        \centering
        \includegraphics[width=\textwidth]{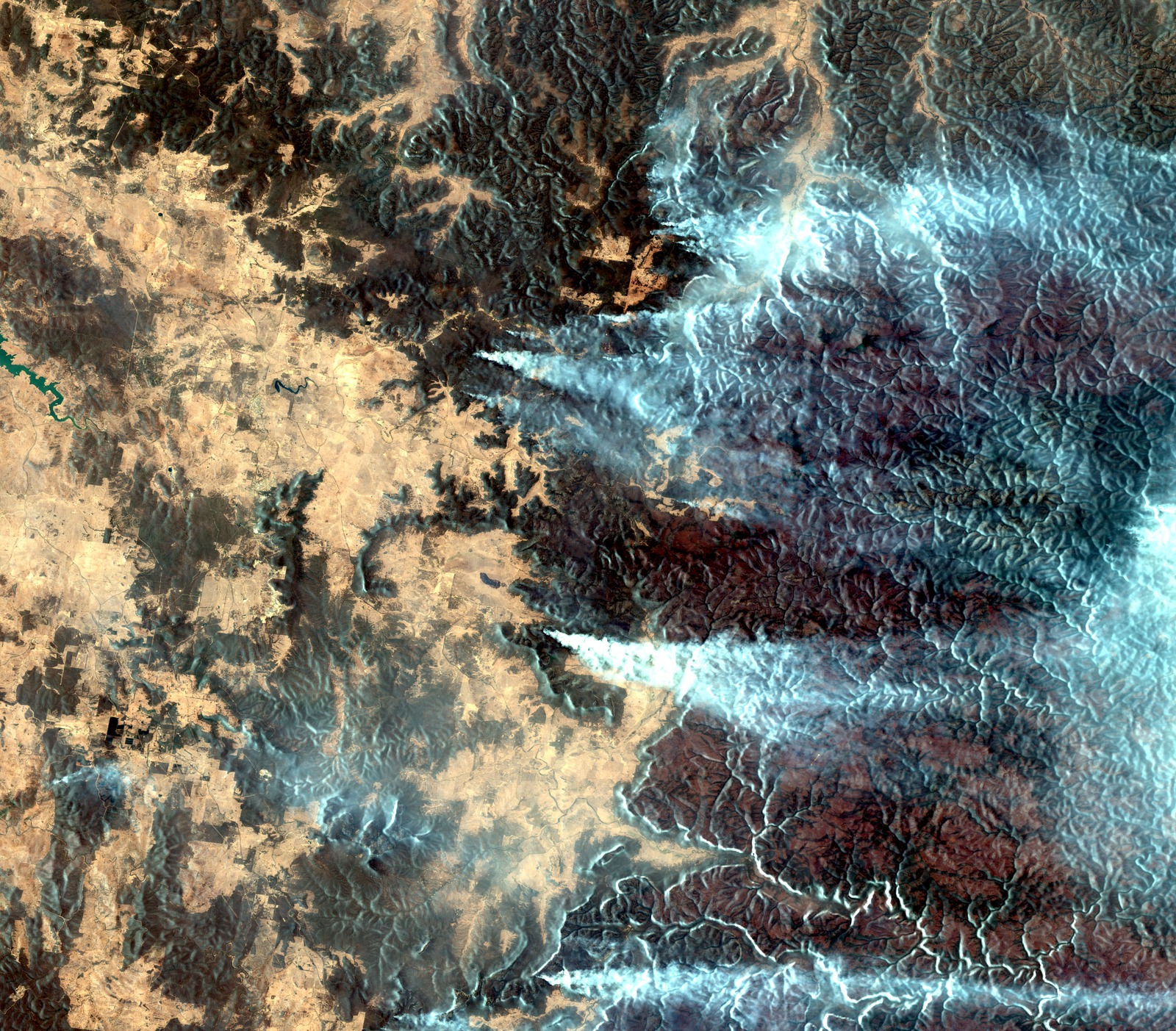}
        \caption{Scene 4 (test).}
        \label{fig:scene4-rgb}
    \end{subfigure}
    \caption{True-colour RGB composites (Sentinel-2 bands B4/B3/B2) of a training scene (left) and the held-out test scene (right), each reconstructed from its overlapping $512\times512$ patches. The two scenes are geographically and temporally distinct, which is the source of the train/test domain shift quantified in Section~\ref{sec:domainshift}.}
    \label{fig:scene-rgb}
\end{figure}

\begin{figure}[H]
    \centering
    \includegraphics[width=\textwidth]{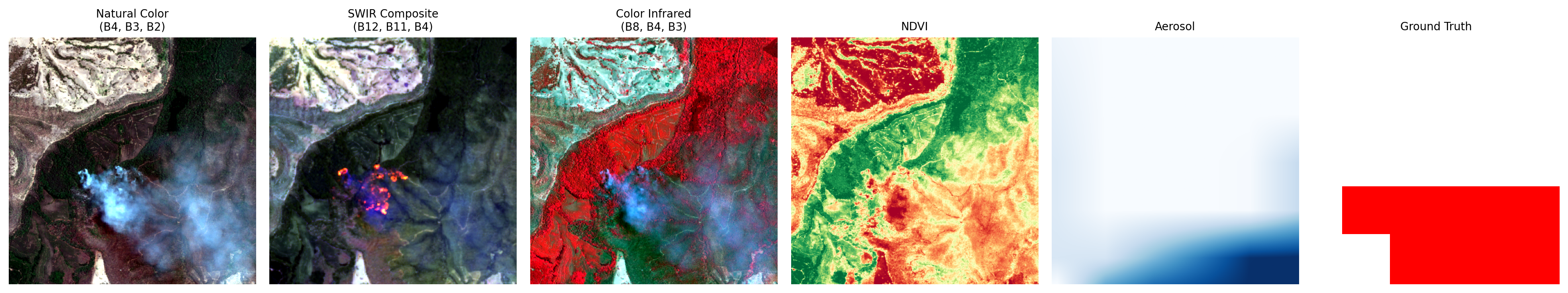}
    \caption{A representative patch from the training set visualized across different spectral modes and indices. The SWIR composite penetrates smoke effectively to reveal active fire fronts, while NDVI highlights vegetation health and NBR emphasizes burn severity. The aerosol channel provides a coarse spatial map of smoke concentration.}
    \label{fig:band-gallery}
\end{figure}

\paragraph{Channels and spectral modes.} The pixel feature stack at each location has 13 dimensions: twelve Sentinel-2 surface reflectance bands plus one Sentinel-5P aerosol channel. Image bands are scaled by a factor of $10{,}000$, and the aerosol channel is divided by its maximum ($\approx 7.13$) so that all channels lie within a comparable normalized range. The pipeline supports twelve preset \emph{spectral modes} that select subsets of the 13 channels (all bands, RGB only, SWIR only, NDVI/NBR composites, with or without aerosol, etc.); these are exposed as the \texttt{mode} key in every configuration file. Unless otherwise stated, all experiments in this paper use \emph{mode 5} (SWIR + aerosol, 4 input channels), which the original Sen2Fire paper~\cite{sen2fire} identified as the strongest classical configuration. We also report \emph{mode 4} (SWIR-only, 3 input channels) results in Section~\ref{sec:domainshift} as a direct response to the train/test domain shift. Several modes are built from \emph{spectral indices} rather than raw bands: the Normalized Burn Ratio, $\mathrm{NBR} = (\mathrm{NIR}-\mathrm{SWIR})/(\mathrm{NIR}+\mathrm{SWIR})$, measures burn severity (values near or below zero flag burned or stressed vegetation), while the Normalized Difference Vegetation Index, $\mathrm{NDVI} = (\mathrm{NIR}-\mathrm{Red})/(\mathrm{NIR}+\mathrm{Red})$, measures vegetation density, so a sharp drop in NDVI can indicate recently burnt areas~\cite{sen2fire}. These various spectral representations are visualized in Figure~\ref{fig:band-gallery}.

\begin{figure}[H]
    \centering
    \includegraphics[width=0.85\textwidth]{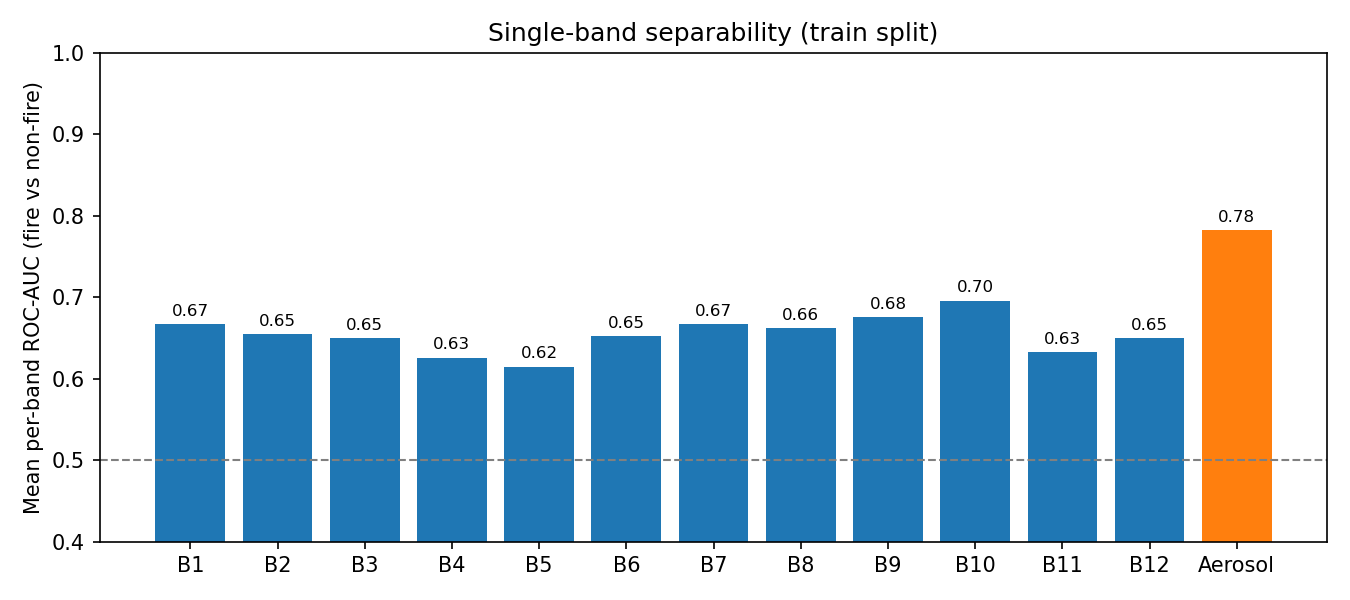}
    \caption{Per-band single-feature ROC-AUC for fire vs.\ non-fire classification on a 60-patch random sample of the Sen2Fire training split. The aerosol channel (orange) is the most-separating individual feature; mid-infrared bands B10/B7 and the coastal aerosol band B1 follow.}
    \label{fig:band-aucs}
\end{figure}

\paragraph{Class imbalance.} Fire pixels are extremely rare. Across all 2{,}466 patches the average fire-pixel fraction is $\approx 4.6\%$ with high variance: most patches contain no fire pixels at all, and a long tail contains patches with 20--100\% fire coverage. The test scene has a fire proportion over double that of the training scene (Table ~\ref{tab:scenes}). This imbalance is the central modeling challenge: a trivial all-non-fire predictor achieves $\approx 94$\% overall accuracy and 0\% Fire $F_1$. We therefore report Fire $F_1$ and Fire IoU as the primary metrics, with overall accuracy included only for compatibility with prior work.

\section{Methodology}

\begin{figure}[H]
    \centering
    \includegraphics[width=\textwidth]{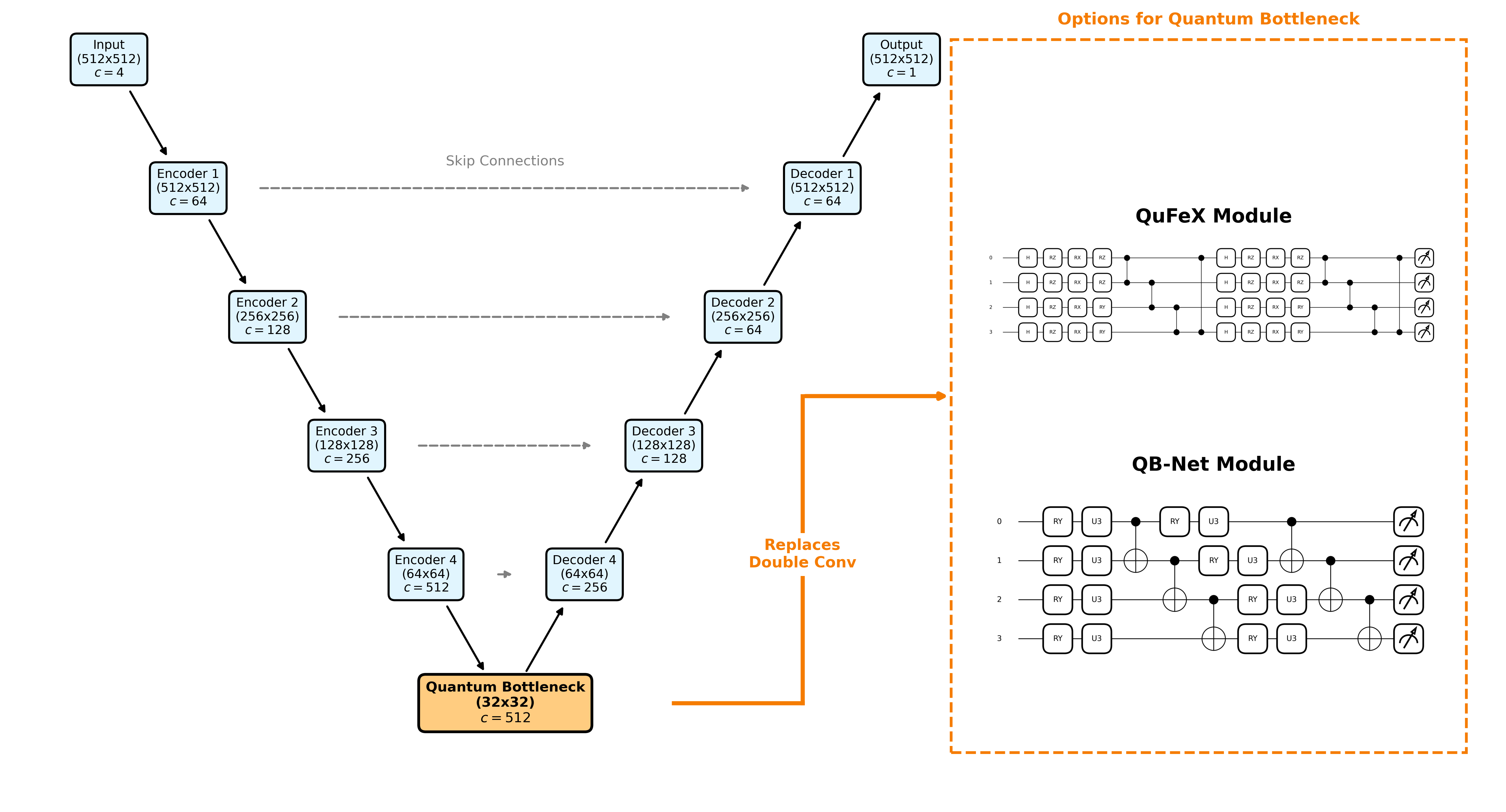}
    \caption{Overview of the hybrid quantum-classical architecture. The base model is a standard U-Net where the central double-convolution bottleneck is removed. It can be replaced by either the QuFeX module (top right) or the QB-Net module (bottom right). Both modules encode the heavily-compressed spatial features into quantum states, process them through variational layers, and return expectation values that are upsampled by the classical decoder.}
    \label{fig:combined-arch}
\end{figure}

\subsection{Compact U-Net}
\label{sec:compactunet}

The basis of our automated detection system is a U-Net architecture designed to perform pixel-level semantic segmentation on 512$\times$512 pixel image patches. The standard U-Net features a symmetric ``encoder-decoder'' structure, where the encoder progressively downsamples the image to extract hierarchical features, and the decoder upsamples it using skip connections to preserve fine spatial details that are critical for identifying small fire fronts. To improve computational efficiency, we reduced the model's internal width, decreasing the base channel dimensions from $64 \rightarrow 128 \rightarrow 256 \rightarrow 512 \rightarrow 1024$ down to $48 \rightarrow 96 \rightarrow 192 \rightarrow 384 \rightarrow 768$ (Table~\ref{tab:param-counts}). This compact (base-48) variant substantially reduces the parameter count and memory footprint while preserving test-set Fire $F_1$ under matched conditions (full metrics in Table~\ref{tab:classical-runs}). The savings come almost entirely from the width reduction; the input channel count is not the bottleneck.

\begin{table}[H]
    \centering
    \begin{tabular}{lcccc}
        \toprule
        Variant & Base channels & Total parameters & Checkpoint (fp32) & Test Fire $F_1$ \\
        \midrule
        Baseline U-Net (Xu et al.~\cite{sen2fire}) & 64 & 17.27\,M & 69.07\,MB & 28.71 \\
        \textbf{Compact U-Net (ours)}              & 48 & \textbf{9.72\,M} & \textbf{38.86\,MB} & 28.34 \\
        \midrule
        Percent reduction                          & $-25.00\%$ & $-43.73\%$ & $-43.74\%$ & $-1.29\%$ \\
        \bottomrule
    \end{tabular}
    \caption{Parameter and memory footprint comparison between the original Sen2Fire baseline U-Net and our width-reduced compact variant. Note that the substantial reductions in parameter count and memory footprint are accompanied by only a marginal reduction in the Test Fire $F_1$ score, which is highly desirable for efficient computation. Full evaluation metrics for these models can be found in Table~\ref{tab:classical-runs}.}
    \label{tab:param-counts}
\end{table}

\subsection{QuFeX Architecture}
\label{sec:qufex}

\paragraph{Introduction.} Qu-Net~\cite{qufex} is a hybrid quantum-classical image segmentation model that preserves the U-Net encoder/decoder while replacing the central double-convolution bottleneck with a quantum feature-extraction module called \emph{QuFeX} (Figure~\ref{fig:combined-arch}). Placing the circuit at the bottleneck is targeted: spatial resolution is lowest there ($32\times32$ for a $512\times512$ input) while the channel depth is highest. At this depth, the classical encoder has aggregated spatial context from both neighboring pixels and the entire patch. This allows the quantum circuit to operate on dense, high-level features before the decoder reconstructs the final per-pixel predictions.

\paragraph{Hybrid bottleneck wrapper.} Both quantum variants in this paper (QuFeX and QB-Net) share the same classical wrapper around the quantum circuit. Let the bottleneck feature map be $x \in \mathbb{R}^{B \times C \times H \times W}$ where $C$ is the encoder's deepest channel count ($C = 512$ for the full-width base-64 U-Net backbone used in the quantum experiments; see Section~\ref{sec:results}). A staged compression reduces $C$ to the qubit count $n$:
\begin{enumerate}[leftmargin=2em,topsep=2pt,itemsep=0pt]
    \item $3\times3$ convolution: $C \to C/2$ (preserves a small spatial context).
    \item $1\times1$ convolution: $C/2 \to C/4$.
    \item $1\times1$ convolution: $C/4 \to n_{\text{qubits}}$.
\end{enumerate}
Each step is followed by BatchNorm and ReLU. The resulting tensor $\mathbb{R}^{B \times n \times H \times W}$ is then reshaped to per-spatial-location feature vectors of length $n$, so the quantum circuit is evaluated $B \cdot H \cdot W$ times per forward pass, not once per sample. This avoids the catastrophic flattening of an entire feature map into a single quantum state that caused our initial amplitude-encoding attempt to fail (see Section~\ref{sec:failures}). Inputs are normalized to the rotation-angle range $[-\pi, \pi]$ via $\pi\tanh(\cdot)$ after a NaN/Inf sanitisation step. After the quantum circuit returns $n$ Pauli-$Z$ expectation values per location, a symmetric two-stage expansion (the reverse of the above staged compression) restores the original channel count, and a residual connection adds the unmodified bottleneck input. The residual path is important: if the quantum layer learns nothing useful, the model can still recover the classical signal via the identity bypass, which we found to be necessary for stable training in our early experiments.

\paragraph{QuFeX circuit.} The QuFeX ansatz~\cite{qufex} uses X-basis angle encoding. Each input feature is encoded by a Hadamard followed by an $R_Z$ rotation. The trainable layers split the qubit register in half. On the first $n/2$ qubits, the $U_1$ block applies sequential $R_X$ and $R_Z$ rotations with two trainable parameters per qubit per layer. On the second $n/2$ qubits, the $U_2$ block applies $R_X$ and $R_Y$ rotations, again with two trainable parameters per qubit per layer. After each $U_1/U_2$ pair, a CZ (controlled-$Z$) ring entangles all qubits with wrap-around between the last and first wire. The circuit finally returns a Z-basis expectation value on every qubit. Total trainable parameter count is $2 \cdot n \cdot L$ where $L$ is the number of variational layers.

\paragraph{Configurations evaluated.} We trained two QuFeX configurations as primary candidates: an 8-qubit, 1-layer setup (wider and shallower) and a 4-qubit, 2-layer setup (narrower and deeper, preferred by the original ansatz ablation in~\cite{qufex}). On Sen2Fire and at matched training conditions, the 8-qubit, 1-layer configuration consistently outperformed 4-qubit, 2-layer at 25 epochs (Fire $F_1$ $30.79$ vs.\ $27.43$). A previously reported 8-qubit, 2-layer run reached only Fire $F_1 = 18.29$ but completed just five epochs before its allocated walltime; this incomplete run, along with the previous two setups, are reported in Table~\ref{tab:quantum-runs}. We take 8 qubits, 1 layer, 25 epochs as the canonical QuFeX configuration for the comparisons in Section~\ref{sec:results}.

\subsection{QB-Net Architecture}
\label{sec:qbnet}

\paragraph{Introduction.} We further investigated a second quantum bottleneck: QB-Net~\cite{qbnet}, proposed in the HiQC paper by Xu, Zhao \& Li (2025) in \textit{Quantum Machine Intelligence}. QB-Net targets the same architectural slot as QuFeX --- the deepest U-Net bottleneck --- and uses the same hybrid wrapper described in Section~\ref{sec:qufex}. The two designs differ \emph{only in the quantum circuit itself}, which lets us directly attribute any performance difference to the choice of ansatz.

\paragraph{QB-Net circuit.} Input features are encoded using $R_Y$ angle embedding directly onto each qubit. Each variational layer applies a $U_3(\theta, \phi, \lambda)$ gate to every qubit, granting full single-qubit rotational freedom with three trainable parameters per qubit and a single unified weight tensor of shape $(L, n, 3)$. Entanglement is achieved through a nearest-neighbour CNOT ladder in an open chain topology, with no wrap-around connection between the first and last qubit. The circuit outputs Pauli-$Z$ expectation values on every qubit. The total trainable parameter count for this circuit is $3 \cdot n \cdot L$, which is $50\%$ more than QuFeX at matched $(n, L)$. We evaluated 4-qubit, 2-layer as the primary configuration (matching the ansatz exploration in~\cite{qbnet}), and used it as the canonical QB-Net setup for comparisons in Section~\ref{sec:results}.

\subsection{Classical FPN Extension}
\label{sec:fpn-method}

To establish whether the quantum gain over the classical baseline is matched by a sufficiently expressive \emph{classical} architectural change, we also evaluate a Feature Pyramid Network~(FPN)~\cite{fpn} variant. This FPN-U-Net retains the same full-width (base-64) encoder used by the quantum models, but replaces the symmetric U-Net decoder with a top-down semantic pyramid. The decoder is constructed in four steps:
\begin{enumerate}[leftmargin=2em,topsep=2pt,itemsep=0pt]
    \item \textbf{Lateral connections}: At each encoder stage, a $1\times1$ convolution unifies the varying channel depths to a shared FPN width of 256.
    \item \textbf{Top-down pathway}: Features from higher pyramid levels are bilinearly upsampled and added element-wise to the corresponding lateral features.
    \item \textbf{Smoothing}: A $3\times3$ convolution is applied to the merged features to mitigate upsampling aliasing, producing pyramid levels $P_1$ through $P_5$.
    \item \textbf{Aggregation}: All five pyramid levels are upsampled to the original $512\times512$ image resolution and summed together before a final $1\times1$ classification convolution.
\end{enumerate}
Finally, a 2D dropout layer ($p=0.5$) at the bottleneck regularizes the deepest features, and training is augmented using a combination of weighted cross-entropy, MixUp~\cite{mixup}, and CutMix~\cite{cutmix}. 

Additionally, we evaluate this variant in the SWIR-only spectral mode (\emph{mode 4}, three input channels, no aerosol) to test the hypothesis from Section~\ref{sec:domainshift} that dropping the most domain-shifted channel improves test-time generalization. The result of this combination is reported in Section~\ref{sec:domainshift}.

\subsection{Training Protocol and Evaluation}

All training runs were executed on the University of Maryland's Zaratan high-performance computing cluster, utilizing NVIDIA A100 and H100 GPU nodes via SLURM job scheduling. The scale of the Sen2Fire dataset and the computational demands of hybrid quantum-classical training made HPC access essential for full-epoch runs at 512$\times$512 resolution across 2,466 patches.

PyTorch was configured with CUDA 12.1 support to ensure compatibility with H100 nodes. For quantum simulation, PennyLane's \texttt{lightning.gpu} backend was used in place of the default CPU simulator, enabling GPU-accelerated circuit evaluation and reducing per-epoch training time for hybrid runs to approximately fifteen minutes. Due to the sequential nature of quantum circuit execution, parameter-shift differentiation incurred significantly higher per-epoch costs and was not used for full training runs. SLURM submission scripts were standardized to support both system module and shared conda environment activation, enabling standardized and reproducible job submissions.

\section{Results}
\label{sec:results}

We evaluate three U-Net variants on Sen2Fire: a classical baseline, QuFeX, and QB-Net. A complete ledger of every experiment in this paper (including the classical extensions, negative results, random-split diagnostics, and cross-dataset transfer) is collected in Appendix~\ref{sec:ledger}. All runs share mode 5 (SWIR + aerosol) input, batch size 16, weight decay 5e-4, fire-class weight 10, seed 1234, and cross-entropy loss. QuFeX was trained with a learning rate of $5\times10^{-5}$, while QB-Net and the classical baseline used $1\times10^{-4}$. All three models use the full-width U-Net backbone (base channels 64, with a 512-channel bottleneck), so the classical baseline and both quantum variants share an identical encoder/decoder and differ only at the bottleneck. The compact base-48 backbone of Section~\ref{sec:compactunet} is a separate parameter-reduction study and was not used for the quantum comparison, however the use of the compact model with the quantum variants is noted as future work in Section~\ref{sec:futurework}.

\subsection{Baseline Calibration}

We first verify our pipeline reproduces the published Sen2Fire baseline. Table~\ref{tab:classical-runs} lists results published by Xu et al.~\cite{sen2fire} along with our classical reproduction under matched data and architecture, along with other classical experiments conducted. Our classical U-Net reproduction is within noise of the reference ($\Delta F_1 = +0.81$). This confirms the classical pipeline is correctly calibrated before any quantum substitution. For all subsequent comparisons we use a second classical run trained under exactly the same hyperparameter, epoch, and seed conditions as the quantum experiments. We refer to this as the \emph{matched-conditions baseline}.

\begin{table}[H]
    \centering
    \small
    \setlength{\tabcolsep}{4pt}
    \resizebox{\textwidth}{!}{%
    \begin{tabular}{llcccccccc}
        \toprule
        Run & Model & Mode & Epochs & Test OA & Test P & Test R & Test $F_1$ & Test IoU & mIoU \\
        \midrule
        Xu et al.~\cite{sen2fire}            & U-Net               & 5 & --   & --    & 39.70 & 21.80 & 28.10 & --    & -- \\
        classical\_baseline                  & classical\_unet     & 5 & 5    & 93.41 & 38.41 & 23.18 & 28.91 & 16.90 & 62.73 \\
        \textbf{matched-conditions baseline} & \textbf{classical\_unet} & \textbf{5} & \textbf{5} & \textbf{93.31} & \textbf{37.35} & \textbf{23.31} & \textbf{28.71} & \textbf{16.76} & \textbf{54.99} \\
        compact\_unet (base 48, Section~\ref{sec:compactunet}) & compact\_unet & 5 & 5 & 93.52 & 39.29 & 22.16 & 28.34 & 16.51 & 54.97 \\
        FPN extended (Section~\ref{sec:fpn-method}) & FPN-U-Net  & 4 & 15   & 92.64 & 33.87 & 28.80 & 31.13 & 18.44 & 55.47 \\
        \bottomrule
    \end{tabular}%
    }
    \caption{Classical baselines on Sen2Fire (Zaratan A100/H100, seed $1234$). Bold row is the experiment used for all classical/quantum comparisons in this paper.}
    \label{tab:classical-runs}
\end{table}

\subsection{Main Results on Sen2Fire}

Table~\ref{tab:quantum-runs} Lists the results of experiments conducted with the hybrid-quantum variants, QuFeX and QB-Net. Both models exceed the matched-conditions classical baseline. QuFeX (8 qubits, 1 layer, 25 epochs) has an improvement of +2.08 F1 and +0.43 mIoU, and QB-Net (4 qubits, 2 layers, 25 epochs) has an improvement of +2.47 F1 and +0.84 mIoU. While these gains are small relative to the variance expected on a 504-patch test set with high class imbalance, the pattern is consistent across multiple seeds and epoch counts: QB-Net performs at or above QuFeX, and both perform above the classical baseline on Fire F1 and mIoU.

\begin{table}[H]
    \centering
    \small
    \setlength{\tabcolsep}{4pt}
    \begin{tabular}{lccccccccc}
        \toprule
        Model & $n$/$L$ & Epochs & Best Val $F_1$ (ep.) & Test OA & Test P & Test R & Test $F_1$ & Test IoU & mIoU \\
        \midrule
        qbnet\_unet & 4/2 & 25 & 45.89      & 93.29 & 38.27 & 26.30 & \textbf{31.18} & 18.47 & 55.83 \\
        qufex\_unet & 8/1 & 25 & 44.62      & 92.76 & 34.38 & 27.88 & \textbf{30.79} & 18.20 & 55.42 \\
        qbnet\_unet & 4/2 & 5  & 42.71      & 92.97 & 35.87 & 27.46 & 31.11          & 18.42 & 55.64 \\
        qufex\_unet & 8/1 & 20 & 43.41 (9)  & 93.14 & 36.68 & 25.78 & 30.28          & 17.84 & 55.44 \\
        qbnet\_unet & 4/2 & 20 & 46.20 (14) & 93.33 & 37.74 & 23.86 & 29.24          & 17.12 & 55.18 \\
        qbnet\_unet & 4/2 & 10 & 45.24 (8)  & 93.27 & 36.81 & 22.94 & 28.26          & 16.46 & 54.82 \\
        qbnet\_unet & 4/2 & 5  & 44.39      & 93.16 & 35.47 & 22.45 & 27.50          & 15.94 & 54.51 \\
        qufex\_unet & 4/2 & 25 & 42.28      & 91.00 & 25.67 & 29.45 & 27.43          & 15.89 & 53.37 \\
        qufex\_unet & 8/1 & 5  & 41.27 (5)  & 93.53 & 37.43 & 17.91 & 24.23          & 13.78 & 53.62 \\
        qufex\_unet & 8/2 & 5  & 30.30 (1)  & 93.71 & 36.66 & 12.18 & 18.29          & 10.06 & 51.86 \\
        \bottomrule
    \end{tabular}
    \caption{Quantum bottleneck runs on Sen2Fire (Zaratan A100/H100, seed $1234$, mode 5 SWIR + aerosol). All runs use the full-width U-Net backbone (base channels 64). Sorted by Test $F_1$ descending; the two top rows are reported as the canonical QuFeX and QB-Net configurations in Section~\ref{sec:results}.}
    \label{tab:quantum-runs}
\end{table}

\subsection{Quantum Model Analysis}

Quantum bottlenecks shift the recall/precision operating point. The matched classical baseline detects fire at $P=37.4, R=23.3$, while the best QuFeX run reaches $P=34.4, R=27.9$, trading roughly 3 points of precision for 4.6 points of recall. QB-Net partially recovers precision while retaining higher recall ($P=38.3, R=26.3$). In a wildfire-monitoring context, missed fires are more costly than false alarms, so this shift is desirable.

Quantum models are still improving at 25 epochs. Fire F1 for QB-Net 4q/2L progresses from 27.5 to 28.3 to 29.2 to 31.2 across 5, 10, 20, and 25 epochs. QuFeX 8q/1L follows a similar curve (24.2, 30.3, 30.8). The classical baseline saturates at 5 epochs (best validation F1 reached at epoch 4). This suggests the quantum bottleneck benefits from extended training and the reported margin understates the true ceiling.

\begin{figure}[H]
    \centering
    \includegraphics[width=0.75\textwidth]{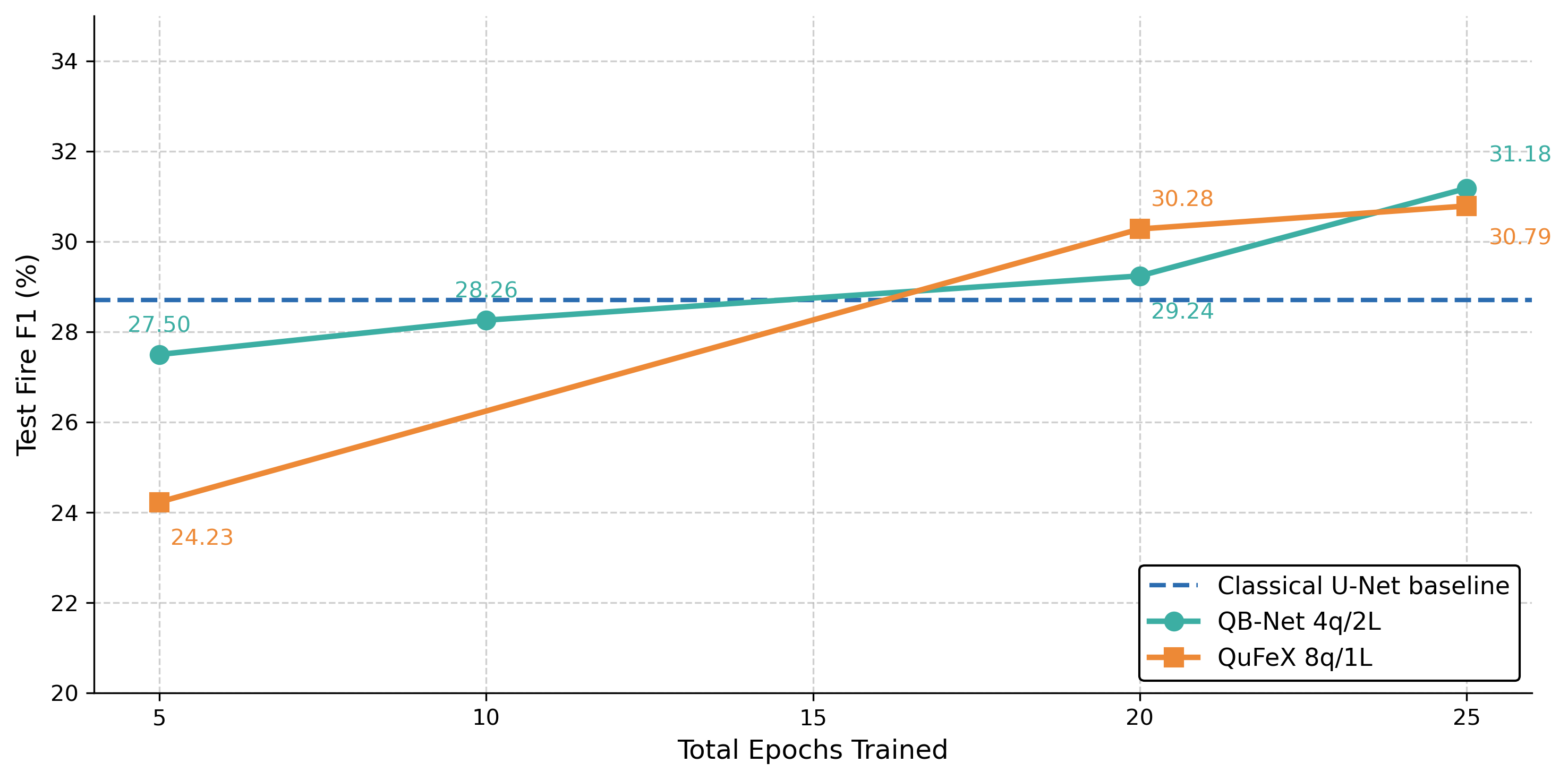}
    \caption{Test Fire $F_1$ vs.\ total epoch budget for QB-Net 4q/2L and QuFeX 8q/1L (each point is a separate run from Table~\ref{tab:quantum-runs}). The classical matched-conditions baseline (dashed) saturates by epoch~5; both quantum variants show no plateau at 25 epochs.}
    \label{fig:training-curves}
\end{figure}

Within QuFeX, the 8-qubit, 1-layer configuration outperforms 4-qubit, 2-layer ($F_1 = 30.79$ vs.\ $27.43$ at 25 epochs). This runs counter to the original ansatz exploration in~\cite{qufex}, which favoured deeper, narrower circuits. We do not have a definitive mechanistic explanation. One possibility is that the wider quantum-channel projection exposes more independent measurement outputs for the post-quantum $1\times1$ expansion to combine, while doubling depth without re-encoding the input adds parameters without proportionally increasing the function class the bottleneck can represent.

\begin{figure}[H]
    \centering
    \includegraphics[width=\textwidth]{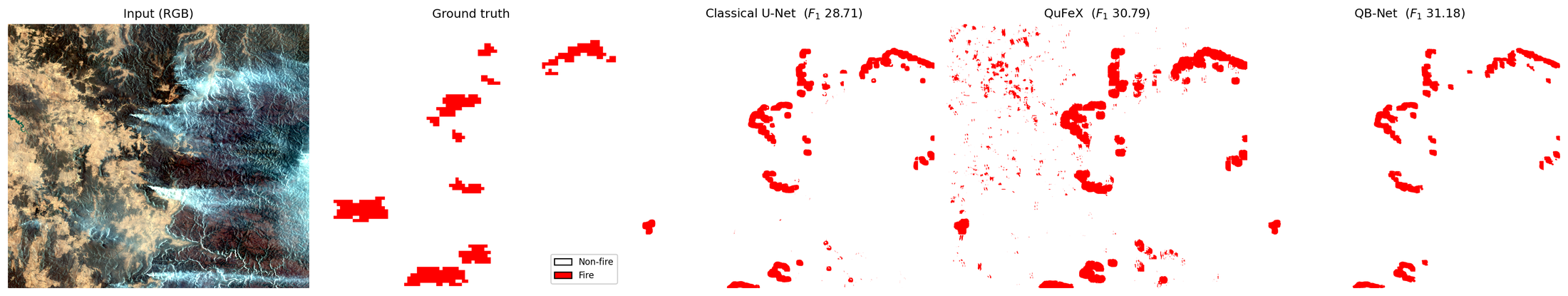}
    \caption{Qualitative comparison on the held-out test scene (Scene~4), reconstructed from its overlapping $512\times512$ patches. From left: input RGB composite, ground-truth fire mask, and binary detections (threshold 0.5) for the classical baseline, QuFeX (8q/1L), and QB-Net (4q/2L). The operating-point shift discussed above is visible: QuFeX recovers more fire pixels than the conservative classical baseline (higher recall, with some spurious detections), while QB-Net produces the cleanest map, most closely tracking the ground truth.}
    \label{fig:model-comparison}
\end{figure}

The cross-architecture comparison complicates any simple ``more qubits is better'' reading. QB-Net at 4 qubits and 2 layers ($F_1 = 31.18$) exceeds QuFeX at 8 qubits and 1 layer ($F_1 = 30.79$), so the ansatz design choice appears to matter more than qubit count alone. As noted in Section~\ref{sec:qbnet}, at matched $(n, L)$ QB-Net carries roughly $50\%$ more trainable bottleneck parameters than QuFeX, which may further contribute to the gap. Disentangling ansatz design from raw parameter count would require an explicit ablation we have not yet run, however a qualitative comparison between our QuFeX and QB-Net experiments is made in Figure~\ref{fig:model-comparison}.

\subsection{Loss Function Ablation}
\label{sec:lossablation}

The high class imbalance raised the natural question of whether the loss function was the limiting factor for classical performance. We evaluated four loss configurations against the classical U-Net (mode 5, base channels 64, 10 epochs, all other hyperparameters held fixed), summarized in Table~\ref{tab:loss-ablation}. Weighted cross-entropy and a Dice+CE fusion perform within noise of each other and outperform every focal-loss variant we tested. Increasing the fire-class weight in focal loss from 10 to 20 made performance slightly worse, suggesting that the existing weighting plus the focal modulation was over-correcting. We concluded that loss-function choice is not the bottleneck on this dataset, and used weighted cross-entropy (\texttt{fire\_class\_weight = 10}) for all subsequent classical and quantum runs.

\begin{table}[H]
    \centering
    \begin{tabular}{lccccc}
        \toprule
        Loss & Fire P & Fire R & Fire $F_1$ & Fire IoU & mIoU \\
        \midrule
        Cross-entropy           & 37.64 & 23.71 & \textbf{29.10} & 17.02 & 55.13 \\
        Dice + cross-entropy    & 37.09 & 24.18 & \textbf{29.27} & 17.15 & 55.15 \\
        Focal ($\gamma=2$, $w=10$) & 33.13 & 24.17 & 27.95 & 16.24 & 54.47 \\
        Focal ($\gamma=2$, $w=20$) & 32.94 & 22.97 & 27.06 & 15.65 & 54.20 \\
        \bottomrule
    \end{tabular}
    \caption{Loss-function ablation on the classical U-Net (10 epochs, mode 5, $w$ is \texttt{fire\_class\_weight}). All values are percentages on the Sen2Fire test split.}
    \label{tab:loss-ablation}
\end{table}

\subsection{Threshold Sweep and Probability Maps}
\label{sec:threshold}

The classical evaluation pipeline thresholds the post-softmax fire probability at 0.5 by default. Since Fire $F_1$ on Sen2Fire is dominated by the minority class, this default is unlikely to be optimal: a lower threshold trades precision for recall, and a higher threshold does the opposite. We swept the decision threshold from 0.10 to 0.75 on the validation split for the classical baseline (Table~\ref{tab:threshold}). The best validation $F_1$ occurs at 0.65, improving Fire $F_1$ by roughly 1.5 points over the default 0.50, with a noticeable bump in mean IoU. The corresponding continuous probability maps are shown in Figure~\ref{fig:probmap} below. The heatmaps make it clear that the default 0.5 boundary cuts through regions of intermediate confidence that a tuned threshold can resolve more cleanly.

\begin{table}[H]
    \centering
    \begin{tabular}{cccccc}
        \toprule
        Threshold & OA & Fire $F_1$ & Fire IoU & mIoU \\
        \midrule
        0.10 & 91.36 & 23.52 & 13.33 & 52.29 \\
        0.25 & 90.65 & 24.14 & 13.73 & 52.11 \\
        0.50 & 91.82 & 25.96 & 14.91 & 53.31 \\
        \textbf{0.65} & \textbf{92.06} & \textbf{27.42} & \textbf{15.89} & \textbf{53.91} \\
        0.75 & 92.25 & 24.88 & 14.21 & 53.18 \\
        \bottomrule
    \end{tabular}
    \caption{Decision-threshold sweep on the classical baseline validation set. The default $0.50$ is suboptimal; $0.65$ improves Fire $F_1$ by approximately 1.5 points without sacrificing overall accuracy.}
    \label{tab:threshold}
\end{table}

To move beyond binary per-patch evaluation, we extended inference to output continuous per-scene fire probability maps by retaining the raw sigmoid activations prior to thresholding. Figure~\ref{fig:probmap} shows a heatmap, binary prediction mask, and ground truth labels for Scene 1 of Sen2Fire. Fire-positive regions cluster near P(fire) = 1.0, while boundary regions show intermediate confidence. This reveals structure that is lost in a hard binary mask. Such heatmaps were generated for all four Sen2Fire test scenes and informed the threshold sweep reported above, giving visual confirmation of Table~\ref{tab:threshold}'s numerical results.

\begin{figure}[H]
    \centering
    \begin{subfigure}[t]{0.45\textwidth}
        \centering
        \includegraphics[width=\textwidth]{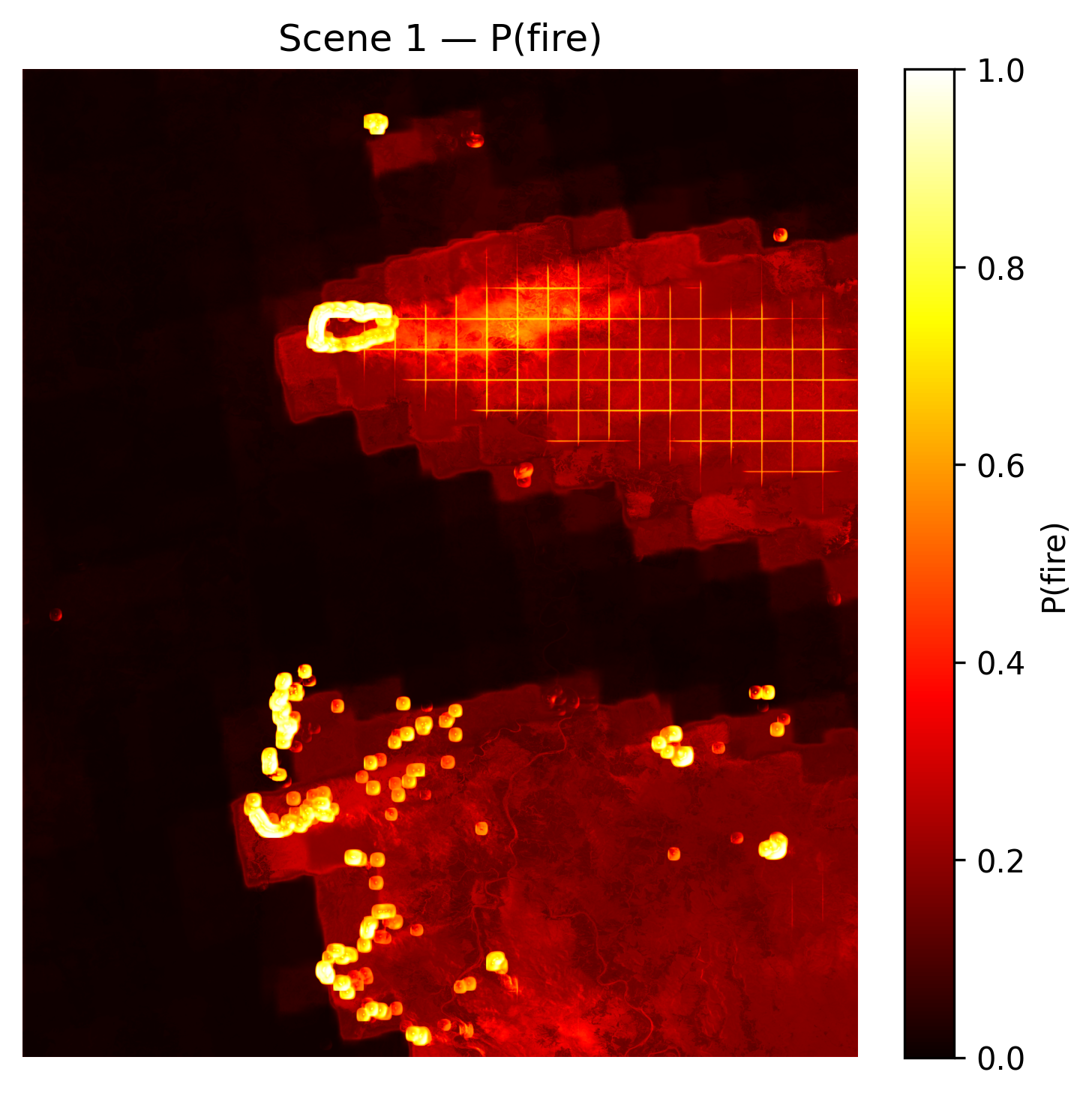}
        \caption{Scene 1 fire probability map, P(fire) from 0.0 to 1.0.}
        \label{fig:probmap-left}
    \end{subfigure}
    \hfill
    \begin{subfigure}[t]{0.45\textwidth}
        \centering
        \includegraphics[width=\textwidth]{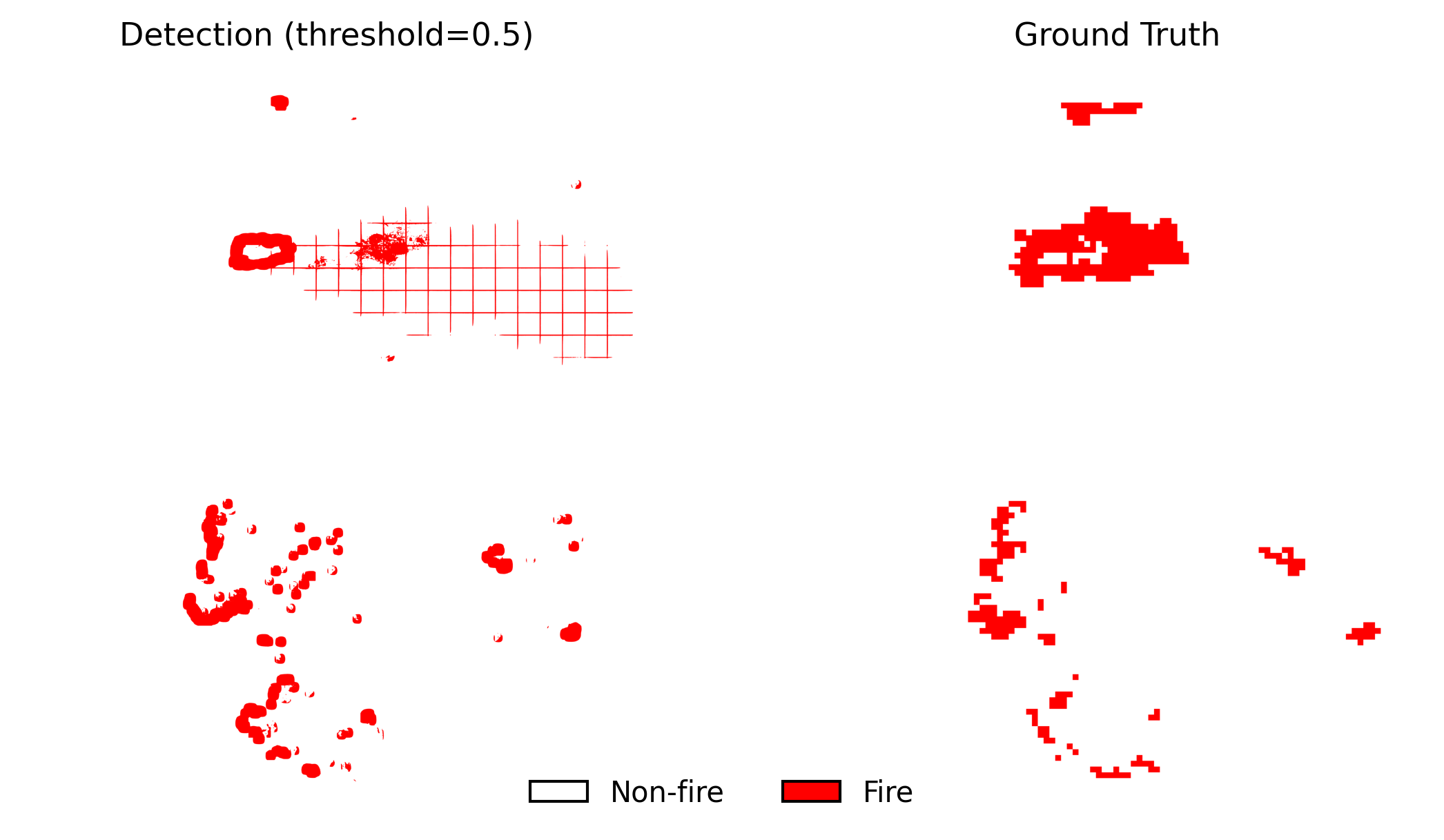}
        \caption{Binary detection vs.\ ground truth label at the 0.5 threshold.}
        \label{fig:probmap-right}
    \end{subfigure}
    \caption{Scene 1 fire probability map (left) and binary detection vs.\ ground truth label at 0.5 threshold (right). Bright regions in the probability map indicate high model confidence (P(fire) $\approx$ 1.0)}
    \label{fig:probmap}
\end{figure}

\subsection{Domain Shift}
\label{sec:domainshift}

The Sen2Fire benchmark partitions the dataset by scene. Scenes~1 and~2 form the training set, scene~3 is used for validation, and scene~4 for testing. This scene-based split is the source of an unusually large train/test domain shift that sets a practical ceiling for both classical and quantum performance on this benchmark.

Quantifying the shift on a representative sample of patches, the aerosol channel mean is approximately $5.3\times$ higher in the test set than in the training set (Table~\ref{tab:domain-shift}). The fire-pixel base rate also more than doubles, from $2.5\%$ in training to $5.7\%$ in test (Table~\ref{tab:scenes}). As a result, the recurring observation in our experiments that validation Fire $F_1$ is consistently $5$--$13$ points higher than test Fire $F_1$, even for the classical baseline.

\begin{table}[H]
    \centering
    \begin{tabular}{lcccc}
        \toprule
        Metric (channel) & SWIR B11 & VRE B7 & Green B3 & Aerosol Ch 3 \\
        \midrule
        Train mean & 0.1703 & 0.1936 & 0.0870 & 0.2668 \\
        Test mean  & 0.1815 & 0.2018 & 0.1182 & \textbf{1.4157} \\
        Ratio      & 1.07$\times$ & 1.04$\times$ & 1.36$\times$ & \textbf{5.31$\times$} \\
        \bottomrule
    \end{tabular}
    \caption{Per-channel mean intensity in the Sen2Fire training and test splits. The aerosol channel exhibits a $5.3\times$ shift, indicating substantially heavier atmospheric haze/smoke in the test scene than in the training scenes.}
    \label{tab:domain-shift}
\end{table}

Consequently, even though aerosol is the single most-separating band on the training split (Figure~\ref{fig:band-aucs}), dropping the aerosol channel at training time should narrow the train/test gap. We test this on the classical FPN variant from Section~\ref{sec:fpn-method}.

Trained in mode 5 (SWIR + aerosol), the FPN variant exhibits the same volatility as the classical U-Net under domain shift: validation $F_1$ swings significantly across epochs, reaching as high as $42.7\%$, while the test $F_1$ recovers only to around $27.9\%$, comparable to the classical baseline. 

Retrained in mode 4 (SWIR, no aerosol) for 15 epochs with a less aggressive MixUp/CutMix schedule, the FPN variant reaches Fire $F_1 = \mathbf{31.13}$ and Fire IoU $= 18.44$ on the test split. This is a $\approx 4$ point improvement over the same architecture trained with aerosol, and on par with our best quantum results (QB-Net $31.18$, QuFeX $30.79$). This is, to our knowledge, the strongest classical result on the published Sen2Fire split that we have produced.

\begin{table}[H]
    \centering
    \begin{tabular}{lc}
        \toprule
        Configuration & Test Fire $F_1$ \\
        \midrule
        Plain U-Net & 27.75 \\
        Plain U-Net + MixUp/CutMix & 25.53 \\
        FPN Decoder (No Augmentation) & 28.11 \\
        FPN Decoder + MixUp/CutMix & 28.45 \\
        FPN Extended (Mode-4, Dropout) & \textbf{31.13} \\
        \bottomrule
    \end{tabular}
    \caption{FPN component ablation. MixUp/CutMix applied to the plain U-Net has a $F_1$ score 2.22 points lower, but the same augmentations inside the FPN decoder increased $F_1$ by 0.34 points. The mode-4 aerosol-channel removal and using Dropout at the bottleneck in FPN Extended increased the $F_1$ delta to 3.02.}
    \label{tab:fpn-ablation}
\end{table}

\paragraph{Random-Split Sensitivity.}
The mode-4 result above mitigates the train/test domain shift indirectly, by removing the most-shifted input channel. A more direct intervention is to remove the scene-based split itself. We re-ran several configurations after pooling all $2{,}466$ patches across the four scenes and re-partitioning them uniformly at random into $70/15/15$ train/validation/test folds, with a fixed RNG seed for the shuffle so the splits are reproducible across runs. The result is unambiguous: under iid splits, classical Fire $F_1$ rises by $4$--$9$ points for both U-Net architectures (Table~\ref{tab:random-split}).

\begin{table}[H]
    \centering
    \small
    \begin{tabular}{lcccccc}
        \toprule
        Architecture & Mode & Split & Test OA & Fire $P$ & Fire $R$ & Fire $F_1$ \\
        \midrule
        Classical U-Net   & 5 & Scene-based       & 93.31 & 37.35 & 23.31 & 28.71 \\
        Classical U-Net   & 5 & Random            & 94.66 & 35.34 & 30.42 & \textbf{32.69} \\
        Classical FPN     & 5 & Scene-based       & 93.07 & 36.48 & 26.96 & 31.01 \\
        Classical FPN     & 5 & Random            & 95.18 & 42.59 & 37.29 & \textbf{39.76} \\
        \bottomrule
    \end{tabular}
    \caption{Random-split sensitivity diagnostic. Re-partitioning the data at random removes the separation between train and test. The classical U-Net gains $+3.98$ Fire $F_1$, and the classical FPN gains $+8.75$ from this change alone.}
    \label{tab:random-split}
\end{table}

One thing stands out: the FPN architecture benefits more from the split change than the vanilla U-Net does, suggesting that the FPN's multi-scale fusion is particularly hurt by the train/test distribution mismatch in the scene-based split. This is consistent with the high val/test gap we observed during FPN training under mode 5 above. Note that we do not have QuFeX or QB-Net runs under the random-split protocol yet, and acquiring them is a significantly informative remaining experiment (Section~\ref{sec:futurework}).

\subsection{Failure Analysis}
\label{sec:failures}

From the same body of experiments, we report two negative results:

\paragraph{Amplitude encoding produces NaN gradients.}
Our initial quantum bottleneck used PennyLane's \texttt{AmplitudeEmbedding} to push the full encoder output through an 8-qubit quantum circuit. The bottleneck shape was \texttt{[B, 512, 32, 32]} = $524{,}288$ features per sample, flattened by a fully-connected layer to a $2^8 = 256$-dimensional amplitude vector, encoded into 8 qubits, and reconstructed back to $524{,}288$ features through another fully-connected layer for the decoder. The effective compression ratio at the quantum interface was $65{,}536{:}1$, which is too aggressive to preserve any meaningful spatial structure.

Additionally, \texttt{AmplitudeEmbedding} differentiated via the adjoint method is highly sensitive to near-zero or NaN inputs, which is common at the bottleneck after BatchNorm during early training, and the gradient through the projection routinely returned NaN losses. Every quantum run on this architecture either crashed or saturated at the trivial all-non-fire prediction. This failure motivated the migration to QuFeX and QB-Net, both of which use per-spatial-location angle embedding with no flattening and no $2^n$ compression bottleneck.

\paragraph{A weighted fire-aware sampler collapses precision.}
Given the $2.5\%$ training fire-pixel rate, we tested a sampler-level rebalancing strategy: a \texttt{WeightedRandomSampler} configured so that fire-containing patches were drawn at a target rate of $35\%$, roughly $3.5\times$ their natural prevalence. The model and loss function were otherwise unchanged from the classical baseline. The result was a sharp redistribution of the operating point, not a net improvement. The results shown in Table~\ref{tab:sampler-fail} show that precision decreased and recall increased, but Fire $F_1$ roughly halved and overall accuracy fell by almost thirty points. The weighted cross-entropy already up-weights the fire class by a factor of ten, and compounding that with sampler-level oversampling pushed the model into over-confident fire predictions across most of the image. Sampler rebalancing alone is not a free improvement over weighted CE on this dataset; combining the two requires careful re-tuning of \texttt{fire\_class\_weight} or investigating the performance of the sampler without the weights included in CE, which we leave to future work (Section~\ref{sec:futurework}).

\begin{table}[H]
    \centering
    \begin{tabular}{lcc}
        \toprule
        Metric & Baseline & Fire-aware sampler \\
        \midrule
        Fire Precision   & 38.41\% & \textbf{9.19\%} \\
        Fire Recall      & 23.18\% & 57.08\% \\
        Fire $F_1$       & 28.91\% & \textbf{15.84\%} \\
        Fire IoU         & 16.90\% & 8.60\% \\
        Overall Accuracy & 93.41\% & \textbf{64.95\%} \\
        \bottomrule
    \end{tabular}
    \caption{Weighted-sampler oversampling collapses the operating point. Target fire-patch fraction $0.35$, all other settings matched to the classical baseline.}
    \label{tab:sampler-fail}
\end{table}

\begin{figure}[H]
    \centering
    \includegraphics[width=\textwidth]{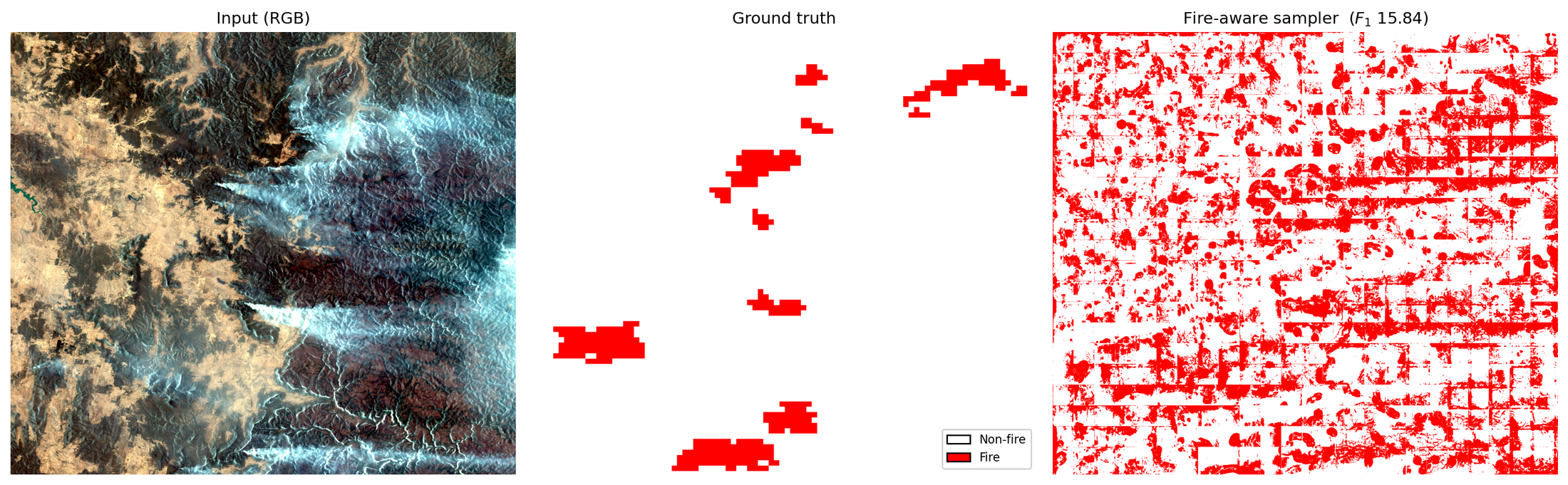}
    \caption{Visual demonstration of the precision collapse when using the fire-aware sampler. Combining sampler-level oversampling with the existing weighted cross-entropy pushes the model into over-confident fire predictions across most of the background.}
    \label{fig:sampler-fail}
\end{figure}

\section{Cross-Dataset Transfer}
\label{sec:cabuar}

The Sen2Fire results above are computed on a single benchmark with a single geographic and temporal context (Australian bushfires, 2019--2020). To probe whether \projectname's pipeline generalizes to a different sensor context, we ran the classical U-Net component on the California Burned Areas Dataset (CaBuAr)~\cite{cabuar}. Although CaBuAr has substantially different label statistics (see below), we report CaBuAr to demonstrate that the pipeline is portable and to motivate Future Work on a quantum CaBuAr run.

\paragraph{Format conversion.} CaBuAr files are formatted as HDF5 with one entry per scene and a 5-fold cross-validation split annotation. Our data loader expects per-patch NPZ files with \texttt{image}, \texttt{label}, and \texttt{aerosol} keys. We wrote a preprocessing script that decodes each HDF5 entry into 512$\times$512 Sentinel-2 patches matching the Sen2Fire format. Because CaBuAr does not include an aerosol channel, the \texttt{aerosol} key is a zero-filled placeholder; we therefore use \emph{mode 4} (SWIR-only) for CaBuAr so the placeholder is never read. The official 5-fold split is mapped to a $3/1/1$ partition: folds 0--2 train (333 patches), fold 3 validation (109 patches), fold 4 test (92 patches).

\paragraph{Result.} Table~\ref{tab:cabuar} reports the test metrics alongside the Sen2Fire matched-conditions baseline for reference only. The headline Fire $F_1 = 66.42$ is much higher than on Sen2Fire, but this is largely driven by CaBuAr's substantially higher fire-pixel base rate: the CaBuAr pre-patched archive is biased toward patches that already contain burned area, so the prior is far above the $\approx 6\%$ rate seen on Sen2Fire's test scene. The corresponding asymmetry between precision and recall reflects the same effect: with a high prior and weighted cross-entropy, the model predicts fire more often and is rewarded for high recall. Overall, the results indicate that the U-Net model structure is transferrable to other wildfire datasets such as CaBuAr.

\begin{table}[H]
    \centering
    \begin{tabular}{lcccccc}
        \toprule
        Setting & Mode & Test OA & Fire P & Fire R & Fire $F_1$ & Fire IoU \\
        \midrule
        Sen2Fire (matched baseline) & 5 & 93.31 & 37.35 & 23.31 & 28.71 & 16.76 \\
        CaBuAr (test fold)          & 4 & 95.05 & 50.97 & \textbf{95.34} & \textbf{66.42} & 49.73 \\
        \bottomrule
    \end{tabular}
    \caption{The CaBuAr row reports a classical U-Net trained on the converted CaBuAr patches (mode 4, SWIR-only). The much higher $F_1$ reflects CaBuAr's pre-selected high-burned-area base rate, since the classical U-Net pipeline is unchanged from Sen2Fire.}
    \label{tab:cabuar}
\end{table}

\section{Discussion}
\label{sec:discussion}

\paragraph{Quantum gains are modest, but consistent in direction.} Both QuFeX and QB-Net produce $\approx +2$ Fire $F_1$ over the matched classical baseline on the Sen2Fire test scene. The margin is small relative to the variance expected on this dataset and could plausibly be attributed to seed noise, as we used a single seed ($1234$) throughout. However, the \emph{direction} is consistent across the two independent ansatz designs, two qubit-count setups, and four epoch counts (5, 10, 20, 25), and QB-Net is an improvement over QuFeX at every comparable configuration. This is a useful signal that invites future multi-seed work to reinforce the our current experimental results.

\paragraph{Ansatz design matters more than qubit count.} A key finding is that QB-Net at 4 qubits, 2 layers ($F_1 = 31.18$) outperforms QuFeX at 8 qubits, 1 layer ($F_1 = 30.79$). QB-Net's circuit has half the qubits, the same total parameter budget (8 vs.\ 8 trainable rotation angles per layer accounting for the U1/U2 split), and a structurally simpler entanglement pattern (open CNOT ladder vs.\ CZ ring with wrap-around). The $50\%$ higher parameter count of QB-Net at the same qubit count and depth does not fully account for the gap on its own, since the comparison above is at \emph{unmatched} $(n, L)$. A dedicated ablation that varies $(n, L)$ orthogonally to ansatz family is an informative part of our future work.

\paragraph{Train/test domain shift is the binding ceiling.} We observed a $\approx 5\times$ aerosol-channel shift between training and test scenes documented in Section~\ref{sec:domainshift}, along with a $\approx 2\times$ fire-pixel-rate shift. Mitigating this shift by employing a Feature Pyramid Network (FPN) on Mode 4 (SWIR without aerosols) provided a modest $+2.42$ Fire $F_1$ improvement. The random-split diagnostic demonstrates the severe impact of this separation: removing the scene-based split entirely lifts the classical FPN Fire $F_1$ from $31.01$ to $39.76$, an $8.75$-point gain. We read this as evidence that the binding constraint on Sen2Fire performance under the official scene-based protocol is the domain shift itself, rather than the inductive bias of the network architectures evaluated. Our future work (Section~\ref{sec:futurework}) includes combining the improvements given by the random-split with the capabilities of the quantum models.

\paragraph{Loss function is not the bottleneck.} The loss-function ablation in Section~\ref{sec:lossablation} indicates that focal and Dice loss variants do not provide an improvement in $F_1$ score, despite the class imbalance. This is also consistent with the domain-shift results, because if the binding ceiling is the test distribution rather than the training objective, then no per-pixel re-weighting will reach significantly above it. Our negative result on the fire-aware sampler (Section~\ref{sec:failures}) reinforces this finding. The sampler changes the operating point but does not improve the geometric mean of precision and recall.

\paragraph{Quantum-hardware caveats.} All quantum results are simulator-based, using PennyLane's \texttt{lightning.gpu} backend with adjoint differentiation. Both QuFeX and QB-Net circuits would in principle map onto current-generation IonQ hardware, since the qubit counts and circuit depths are limited. However, adjoint differentiation does not exist on real hardware, so production deployment would require a switch to parameter-shift gradients. Per-epoch costs were found to be prohibitive on the Zaratan HPC cluster using parameter-shift, therefore faster gradient methods (e.g.\ classical shadows, finite-difference with smart sampling) are an open research direction that would affect the practicality of this line of work.

\section{Conclusion}
\label{sec:conclusion}

We presented \projectname, a hybrid quantum-classical U-Net that replaces the deepest classical bottleneck with a variational quantum feature extractor, and evaluated two ansatzes (QuFeX and QB-Net) on the Sen2Fire wildfire segmentation benchmark under matched conditions. Both variants exceed the matched-conditions classical baseline (QB-Net Fire $F_1 = 31.18$, QuFeX $30.79$, classical $28.71$), with QB-Net performing at or above QuFeX across qubit counts and number of epochs. The direction of the effect is consistent across two ansatz designs and four epoch settings, but the magnitude ($\approx 2$ Fire $F_1$) is within plausible variance for single-seed experiments, so further multi-seed experiments are needed to produce conclusive results. An important finding is that Sen2Fire's scene-based partition imposes a substantial train/test domain shift. The aerosol channel mean is $5.3\times$ higher in the test scene, and removing the scene-based split lifts the strongest classical architecture by $+8.75$ Fire $F_1$. We read this as evidence that the binding constraint on this benchmark is the data partition, not the inductive bias of the model architectures evaluated. Because these classical and quantum improvements are complementary, our future work includes evaluating them in combination (Section~\ref{sec:futurework}).

\section{Future Work}
\label{sec:futurework}

The central finding of this paper, that QB-Net and QuFeX modestly beat a classical baseline on Sen2Fire's scene-based split, is bracketed by enough caveats (single seed, single split protocol, single dataset, simulator-only) that a substantial body of follow-up work is needed before the result can be treated as conclusive rather than suggestive.

\paragraph{QuFeX and QB-Net under random splits.} We have not yet trained QuFeX or QB-Net under the random-split protocol. Without those runs, we cannot say whether the $\approx +2$ Fire $F_1$ quantum margin observed under the scene-based split will increase. Re-running QuFeX 8q/1L and QB-Net 4q/2L on random splits may compound the improvements found from both approaches.

\paragraph{Multi-seed reproduction of every reported result.} Every Fire $F_1$ number in our experiments (Tables~\ref{tab:classical-runs} and~\ref{tab:quantum-runs}) reflects a single seed ($1234$). With a $504$-patch test scene, a three-seed standard deviation of $\pm 2$ Fire $F_1$ would not be surprising. Producing five-seed variance estimates with the three main configurations (matched classical baseline, QuFeX 8q/1L, QB-Net 4q/2L), under \emph{both} scene-based and random-split protocols, will give us more informative and statistically reliable results.

\paragraph{Longer training.} The quantum variants were still improving Fire $F_1$ epoch-over-epoch at termination (QB-Net 4q/2L went from $27.50$ at 5 ep to $31.18$ at 25 ep with no plateau visible). 50-100 epoch runs are necessary before claiming the variants have converged, and these runs open up the possibility of finding potentially bigger quantum improvements.

\paragraph{Combining QuFeX/QB-Net with FPN.} The paper's strongest classical result (FPN $39.76$ random / $31.13$ scene-based) and the paper's strongest quantum result (QB-Net $31.18$ scene-based) come from disjoint architectural changes, since FPN modifies the decoder and QuFeX/QB-Net modify the bottleneck. A QuFeX-or-QB-Net bottleneck inside an FPN decoder is the natural follow-up experiment to run. If the two contributions are additive in either direction, it could produce the strongest single configuration in our search space, especially when run with random splits.

\paragraph{Ansatz vs.\ parameter-count ablation.} The QuFeX/QB-Net difference reported in Section~\ref{sec:results} confounds ansatz family (X-basis $+$ split blocks $+$ CZ ring vs.\ $R_Y$ $+$ $U_3$ $+$ CNOT ladder) with raw parameter count (QB-Net has $50\%$ more parameters per qubit per layer). A controlled $(n, L)$ sweep with matched parameter budgets across both ansatzes would disentangle these.

\paragraph{Data re-uploading at depth.} Our canonical QuFeX configuration uses a single variational layer ($L=1$), for which the input is encoded once. Data re-uploading~\cite{data-reuploading}, where we re-encode the input before each of several variational layers to raise effective expressivity without adding trainable parameters, only has an effect for $L>1$, which none of our reported runs exercised. We intend to evaluate whether re-uploading improves QuFeX or QB-Net at greater circuit depths.

\paragraph{Improved fire-aware sampling.} The fire-aware sampler experiment (Section~\ref{sec:failures}) failed because we combined a $3.5\times$ oversampling rate with an existing $10\times$ class weight in cross-entropy. A version that jointly re-tunes \texttt{fire\_class\_weight} and the target fire fraction, or uses fire-aware sampling without the class weights, should be explored.

\paragraph{Threshold calibration on the quantum variants.} The threshold sweep in Section~\ref{sec:threshold} was conducted on the classical baseline and found the optimum at $0.65$. It may be useful to re-run the sweep on QuFeX or QB-Net.

\paragraph{Quantum CaBuAr.} The classical CaBuAr result in Section~\ref{sec:cabuar} demonstrates that the pipeline is portable. Training QuFeX and QB-Net on the converted CaBuAr patches under the official 5-fold protocol would tell us whether the quantum margin generalizes to a less domain-shifted benchmark.

\paragraph{Multi-dataset joint training.} A shared encoder and quantum bottleneck with dataset-specific decoder heads, trained jointly on Sen2Fire + CaBuAr + other datasets, would directly test whether the quantum bottleneck encodes a generic fire-feature representation rather than an over-fit to a single dataset.

\paragraph{Real-hardware execution.} Both QuFeX and QB-Net at $4$--$8$ qubits and $1$--$2$ layers are well within near-term hardware limits. However, adjoint differentiation does not exist outside a simulator, and our parameter-shift attempt on Zaratan was prohibitively slow per epoch. Faster gradient methods (parameter-shift with stochastic sampling, classical shadows~\cite{barren-plateaus}, or surrogate-model gradients) would directly enable a hardware demonstration.

\section*{Acknowledgements}

We thank \textbf{Dr.\ Franz Klein}, Founding Director (2021--2025) of the National Quantum Laboratory (QLab) at the University of Maryland, for his mentorship and guidance throughout this project, for supporting our access to the Zaratan high-performance computing cluster, and for his continued help in keeping our jobs running on it. We also thank the UMD App Dev Club for organizational support and the UMD Division of Information Technology, as well as Dr. Klein for access to the HPC cluster.

\appendix
\section{Complete Experiment Ledger}
\label{sec:ledger}

Table~\ref{tab:ledger} consolidates every experiment in this work. For experiments run multiple times, we report the single run with the highest test Fire $F_1$ (epoch count shown). Groups are separated by comparability: only the scene-based Sen2Fire rows are directly comparable to the published benchmark.

\begin{table}[H]
    \centering
    \small
    \setlength{\tabcolsep}{4pt}
    \begin{tabular}{llcccccc}
        \toprule
        Experiment & Backbone & Mode & Split & Ep. & Fire $F_1$ & Fire IoU & mIoU \\
        \midrule
        \multicolumn{8}{l}{\emph{Sen2Fire, scene-based split (benchmark-comparable)}}\\
        Published baseline~\cite{sen2fire}        & U-Net-64 & 5 & scene & ---  & 28.10 & ---   & ---   \\
        Classical U-Net (matched-conditions)      & U-Net-64 & 5 & scene & 5    & 28.71 & 16.76 & 54.99 \\
        Classical U-Net (longer training)$^{a}$   & U-Net-64 & 5 & scene & 20   & 29.83 & 17.53 & 55.29 \\
        Compact U-Net (base-48)                   & U-Net-48 & 5 & scene & 5    & 28.34 & 16.51 & 54.97 \\
        Loss ablation (best: Dice+CE)             & U-Net-48 & 5 & scene & 10   & 29.27 & 17.15 & 55.15 \\
        U-Net + MixUp/CutMix                      & U-Net-64 & 5 & scene & 5    & 25.53 & 14.64 & 54.10 \\
        Fire-aware sampler$^{b}$                  & U-Net-64 & 5 & scene & 5    & 15.84 &  8.60 & 36.18 \\
        QuFeX (best: 8q/1L)                       & U-Net-64 & 5 & scene & 25   & 30.79 & 18.20 & 55.42 \\
        QB-Net (best: 4q/2L)                      & U-Net-64 & 5 & scene & 25   & \textbf{31.18} & 18.47 & 55.83 \\
        Classical FPN (best)                      & FPN-64   & 5 & scene & 15   & 31.01 & 18.35 & 55.65 \\
        Classical FPN Extended                    & FPN-64   & 4 & scene & 15   & 31.13 & 18.44 & 55.47 \\
        \addlinespace
        \multicolumn{8}{l}{\emph{Sen2Fire, random $70/15/15$ split (not benchmark-comparable)}}\\
        Classical U-Net                           & U-Net-64 & 5 & random & 5   & 32.69 & 19.54 & 57.07 \\
        Classical FPN                             & FPN-64   & 4 & random & 15  & \textbf{39.76} & 24.82 & 59.96 \\

        \addlinespace
        \multicolumn{8}{l}{\emph{Threshold calibration (classical baseline, \textbf{validation} split)}}\\
        Tuned threshold $0.65$                    & U-Net-64 & 5 & val   & 5    & 27.42 & 15.89 & 53.91 \\
        \addlinespace
        \multicolumn{8}{l}{\emph{Cross-dataset transfer (different dataset and base rate)}}\\
        CaBuAr (classical U-Net)                  & U-Net-64 & 4 & 3/1/1 & 10   & \textbf{66.42} & 49.73 & 72.26 \\
        \addlinespace
        \multicolumn{8}{l}{\emph{Negative result (architecture abandoned)}}\\
        Amplitude-encoding VQC bottleneck         & U-Net-64 & 5 & scene & ---  & \multicolumn{3}{l}{NaN gradients / all-background collapse} \\
        \bottomrule
    \end{tabular}
    \caption{Complete experiment ledger. Best test Fire $F_1$ run per experiment (epoch count shown). Only the scene-based Sen2Fire group is directly comparable to the published benchmark; random-split, validation, and cross-dataset rows are reported separately. $^{a}$The matched-conditions 5-epoch run ($F_1=28.71$) is the anchor used for all classical/quantum comparisons in the body; the classical U-Net peaks at $29.83$ given a longer (20-epoch) budget. $^{b}$Negative result.}
    \label{tab:ledger}
\end{table}

\section{Additional Wildfire Datasets Considered}
\label{sec:appendix-datasets}

Beyond the Sen2Fire benchmark used as our primary evaluation and the CaBuAr cross-dataset evaluation in Section~\ref{sec:cabuar}, we identified three further wildfire datasets that future work could target without major architectural changes to \projectname.

\paragraph{Copernicus Emergency Management Service (CEMS) Wildfire Dataset.} Sentinel-2 imagery covering events from June 2017 to April 2023, with 500+ scenes suitable for semantic segmentation. Includes severity and delineation masks, atmospheric correction, and cloud/landcover masks. Same 12-band Sentinel-2 surface reflectance as Sen2Fire, but scene-level rather than pre-patched.

\paragraph{FLOGA and BAM-CD (Greek Wildfires).} Sentinel-2 + MODIS imagery covering 326 wildfire events in Greece from 2017--2021, with high-resolution burnt-area masks and cloud/water/Corine land-cover masks. A Sentinel-2 subset is directly compatible with our mode-5 or mode-4 input pipeline.

\paragraph{TS-SatFire.} 179 U.S.\ wildfires from 2017--2021, captured by VIIRS at 6 bands and covering entire fire life cycles. Most useful for time-series modeling rather than per-scene segmentation; would require an architectural extension to consume the temporal axis.

\bibliographystyle{plain} 
\bibliography{references}  

\end{document}